\documentclass{article}

\usepackage{microtype}
\usepackage{graphicx}
\usepackage{subcaption}
\usepackage{booktabs} 

\usepackage{hyperref}

\usepackage[accepted]{icml2026}

\usepackage{amsmath}
\usepackage{amssymb}
\usepackage{mathtools}
\usepackage{amsthm}
\usepackage{graphicx}
\usepackage{subcaption}
\usepackage{booktabs}
\usepackage{colortbl}
\usepackage{xcolor}
\usepackage{array}
\definecolor{headerblue}{RGB}{219, 234, 254}
\definecolor{accentblue}{RGB}{20, 100, 145}
\usepackage{enumitem}

\usepackage[capitalize,noabbrev]{cleveref}

\theoremstyle{plain}

\theoremstyle{definition}

\theoremstyle{remark}

\usepackage[textsize=tiny]{todonotes}

\icmltitlerunning{The Hallucination Snowball: Error Propagation in Multi-Agent LLM Pipelines}

\begin{document}

\twocolumn[
  \icmltitle{The Hallucination Snowball: Modeling Error Propagation as \\ State Transitions in Multi-Agent LLM Pipelines}



  \icmlsetsymbol{equal}{*}

  \begin{icmlauthorlist}
    \icmlauthor{Prabhjot Singh}{ut,redi}
  \icmlauthor{Bhushan Pawar}{redi}
\end{icmlauthorlist}

\icmlaffiliation{ut}{The University of Texas at Austin, Austin, TX, USA}
\icmlaffiliation{redi}{RediMinds Inc., USA}

\icmlcorrespondingauthor{Prabhjot Singh}{prabhjot.singh@utexas.edu}
\icmlcorrespondingauthor{Bhushan Pawar}{bhushan.pawar@rediminds.com}

  \icmlkeywords{multi-agent LLM pipelines, hallucination propagation, error compounding, agentic AI reliability, retrieval-augmented verification, Markov error model, pipeline verification, factual consistency}

  \vskip 0.3in
]



\printAffiliationsAndNotice{}  

\begin{abstract}
Sequential multi-agent LLM pipelines chain specialized agents without verification at handoffs, creating a structural flaw with measurable and severe consequences. We show that hallucinations injected at Stage 1 do not merely persist; they transform: raw numerical facts become derived computations, then narrative prose, then editorially approved conclusions. At each transformation, detectability degrades near-irreversibly. We formalize this as the \textit{hallucination snowball effect}, a first-order Markov process over four states (\textit{Raw Fact} $\to$ \textit{Derived} $\to$ \textit{Narrative} $\to$ \textit{Invisible}) with empirically measured per-boundary escape probabilities of 24.6\%, 48.3\%, and 89.3\%. Across 346 automatically injected hallucinations in a 4-agent financial analysis pipeline on FinanceBench, gpt-4o detection drops from 72.0\% at Stage 1 to 50.9\%  at Stage 4, and 23.7\% of hallucinations survive completely undetected  in the final output. Even the strongest model tested (Qwen3.5-397B-A17B,  87.0\% at Stage 1) faces a structural ceiling; projected Stage 4 detection is only ${\sim}$60--65\%. Critically, boundary gates using identical RAG verification tools reduce hallucination survival from 58.4\% to 16.2\% versus end-of-pipeline checking (Cohen's $h = -0.911$, $p < 0.000001$), while end-checking alone achieves merely 2.3~pp improvement over no verification. \textit{When  you verify matters more than whether you verify.} Our model predicts  survival for $n$-agent linear pipelines and prescribes  optimal verification resource allocation: invest at $S_1{\to}S_2$ first, where 75.4\% of hallucinations are still catchable, not at $S_3{\to}S_4$ where 89.3\% have already escaped. 
\end{abstract}

\section{Introduction}

Multi-agent LLM pipelines have become the dominant architecture for  complex, high-stakes AI tasks. Frameworks like LangGraph~\cite{langgraph}, AutoGen~\cite{autogen}, and CrewAI~\cite{crewai} make it trivial to chain specialized agents sequentially: Researcher $\to$ Analyst $\to$ Writer $\to$ Reviewer. Each agent receives only its predecessor's text output with no provenance metadata, no confidence scores, and no access to source documents. Every agent trusts its predecessor completely. This architectural choice has a measurable and surprisingly severe consequence.

Consider a hallucination introduced at Stage 1: ``COGS was \$71.2B'' (true value: \$63.1B). By Stage 2, the Analyst computes ``an 8.3\% YoY increase,'' a derived claim built on a fabricated base. By Stage 3, the Writer produces ``Boeing faced significant cost headwinds, with COGS rising 8.3\% to \$71.2B, reflecting supply chain disruptions.'' By Stage 4, the Reviewer approves it as ``well-structured analysis with sound reasoning,'' having been given no source documents and only able to check internal consistency. The hallucination has been laundered completely.

Prior work frames hallucination mitigation as finding a better  detector~\cite{min2023factscore,dhuliawala2023cove,song2024veriscore,wei2024long}. We show this framing  is fundamentally incomplete for sequential multi-agent systems. gpt-4o  detection drops from 72.0\% at Stage 1 to 50.9\% at Stage 4, a 21.1~pp decay regardless of detector sophistication. End-of-pipeline verification achieves only 2.3~pp improvement over no verification (from 60.7\% to 58.4\% survival) because the information required to verify the original claim is structurally destroyed by upstream transformations. The correct framing is propagation control, not detection.

\begin{figure}[h]
  \centering
  \includegraphics[width=\columnwidth]{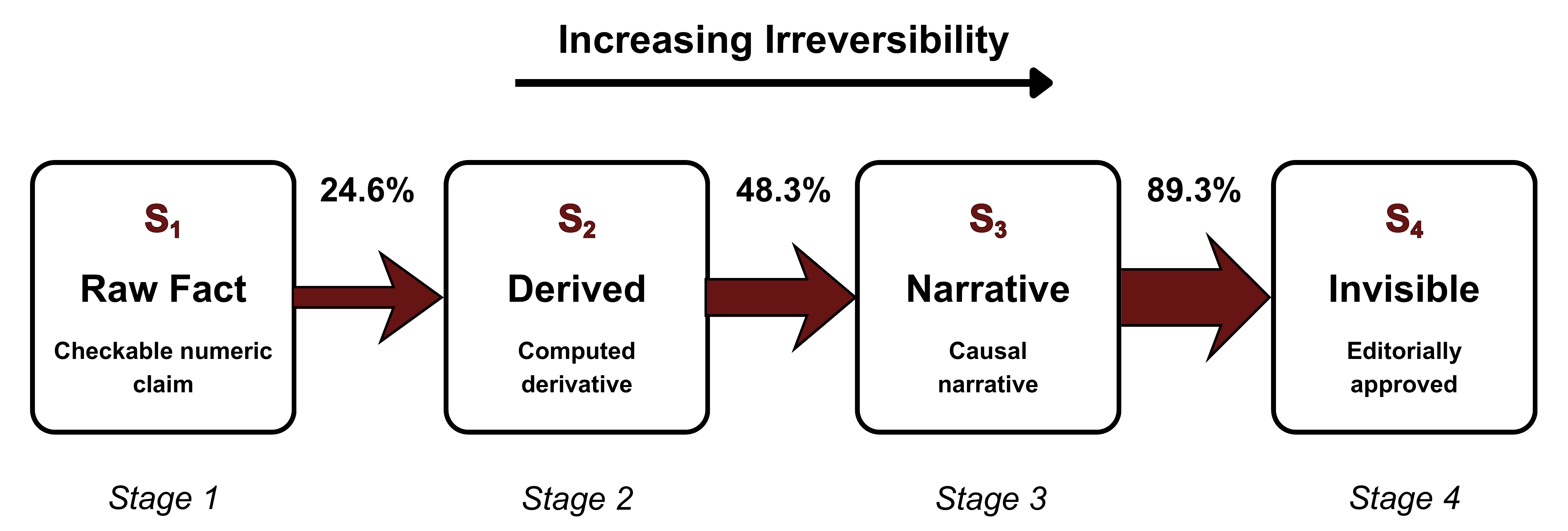}
    \caption{Four-state Markov process of hallucination propagation. Escape probabilities rise from 24.6\% to 89.3\% across boundaries; by $S_3{\to}S_4$, nearly 90\% of hallucinations are structurally unrecoverable by any downstream gate.}
  \label{fig:state_transition}
\end{figure}

We present three experiments and a theoretical model.\renewcommand{\thefootnote}{\fnsymbol{footnote}}\setcounter{footnote}{1}\footnote{Code and results available at \url{https://github.com/prabhjotschugh/hallucination-snowball}.}\renewcommand{\thefootnote}{\arabic{footnote}} Section~\ref{sec:exp1} measures detectability decay across stages. Section~\ref{sec:exp2} characterizes the structural ceiling of LLM skepticism. Section~\ref{sec:exp3} demonstrates that boundary gates reduce survival from 58.4\% to 16.2\%, a 42.2~pp improvement over end-checking (Cohen's $h = -0.911$, $p < 0.000001$), at a measured quality cost of 4.44 to 3.93 on internal consistency. Section~\ref{sec:model} formalizes propagation as a Markov process with measured transition probabilities and prescribes optimal gate placement. Figure~\ref{fig:state_transition} illustrates the four states and their empirically measured escape probabilities. While we demonstrate this in financial analysis, chosen for its unambiguous numeric ground truth, the transformation mechanism is domain-agnostic: any pipeline where a checkable fact can be converted into confident narrative faces identical dynamics. Medical summarization pipelines that chain extraction and synthesis, legal review pipelines that chain precedent retrieval and brief drafting, and research automation pipelines that chain literature search and report generation all share the same structural flaw. The measurement infrastructure and gate architecture we provide are structurally applicable across all of them, though domain-specific escape rates will require calibration outside the numeric-finance setting.

\section{Background and Related Work}

\subsection{Single-Agent Hallucination Detection}
The dominant paradigm treats hallucination as a property of isolated model outputs. FActScore~\cite{min2023factscore} decomposes generations into atomic claims and verifies each against a knowledge source. CoVe~\cite{dhuliawala2023cove} uses self-verification chains where a model checks its own outputs. VeriScore~\cite{song2024veriscore} extends this to long-form generation. SAFE~\cite{wei2024long} uses search-augmented fact evaluation. All of these operate at what we call $S_1$ only: a single output evaluated in isolation. None models what happens when one agent's hallucinated output becomes the next agent's input. Our work shows that a hallucination detectable at 72.0\% at $S_1$ becomes catchable at only 50.9\% three transformations later, making single-agent evaluation an insufficient framework for sequential pipelines.

\subsection{Multi-Agent Debate and Collaboration}
Du et al.~\cite{du2024improving} show that multiple LLMs debating over the same question improves reasoning. MAD~\cite{liang2023encouraging} extends this to structured adversarial debate. Critically, both use parallel topologies where all agents share the same factual foundation simultaneously. Sequential propagation, where each agent sees only the previous agent's transformed output, is structurally absent from this line of work. Our failure mode is orthogonal and addresses the architecture actually deployed in production via LangGraph~\cite{langgraph}, AutoGen~\cite{autogen}, and CrewAI~\cite{crewai}.

\subsection{Trust and Security in Multi-Agent Systems}
TrustAgent~\cite{hua2024trustagent} and work on rogue agents~\cite{yang2024watch} study intentional adversarial manipulation by malicious agents in multi-agent settings. Our failure mode is categorically different: it requires no adversary. Completely well-intentioned agents following their instructions faithfully produce the hallucination snowball because the architecture provides no error-correction mechanism at handoffs. This makes the snowball more pervasive than adversarial threats, which require compromised intent.

\subsection{RAG and Retrieval-Augmented Verification}
Lewis et al.~\cite{lewis2020rag} introduced retrieval-augmented generation to ground LLM outputs in external documents. Subsequent work~\cite{gao2023retrieval} has extended RAG to diverse settings. Our boundary gate is inspired by this line of work but deployed at a fundamentally different layer: not to augment generation, but to verify agent output at handoff points before the next agent processes it. Crucially, our gate is a deterministic numeric matcher (zero LLM calls) rather than a retrieval system; we retain the ``RAG'' label only to signal the grounding intent, not the mechanism. The placement principle, verify before handoff, not before generation, is novel.

\subsection{Concurrent and Complementary Work}
AgentHallu~\cite{agenthallu} independently confirms that early-stage errors dominate downstream failures; where it diagnoses which agent caused a hallucination post-hoc, we measure how detectability degrades across transformations and intervene before transformation occurs. VERIMAP~\cite{verimap} operationalizes per-subtask verification in DAG-structured workflows; our Markov model provides data-driven guidance for where gates yield the highest return within such frameworks. CaveAgent~\cite{caveagent} reduces laundering by avoiding lossy text handoffs entirely, an orthogonal mitigation that trades architectural flexibility for verifiability. Comparing boundary gates against structured-handoff baselines remains an important open question.

\subsection{The Gap We Fill}
No prior work provides a mathematical framework for hallucination transformation across sequential agents, measures where detectability decays most sharply within a pipeline, or demonstrates empirically that verification timing dominates verification tooling. The FMAI workshop calls for trace-level diagnostics and explicit evidence about what improves and what does not. Prior work offers neither for sequential multi-agent systems. We address all three.

\section{Experiments}

\subsection{Setup and Injection Protocol}

\paragraph{Pipeline.} We implement a 4-agent sequential pipeline in LangGraph~\cite{langgraph}: 

\[
\text{Researcher} \to \text{Analyst} \to \text{Writer} \to \text{Reviewer}
\]

All agents use gpt-4o-mini (temperature=0.3) with role-specific system prompts. The Researcher extracts exact figures from SEC filings. The Analyst computes YoY changes and ratios using only Researcher output. The Writer produces a 300--500 word professional narrative. The Reviewer performs internal consistency checking only, with no access to source documents.

\paragraph{Dataset.} We evaluate on FinanceBench~\cite{patronus2023financebench}, 150 expert-annotated financial QA pairs from real SEC filings with exact numeric ground truth. We exclude 10 questions producing qualitative outputs with no injectable numeric values, yielding 140 usable questions. All questions are converted to complex multi-part analytical directives via gpt-4o-mini to reflect realistic pipeline inputs.

\paragraph{Injection Protocol.} We automatically inject 346 hallucinations across 140 questions (2--3 per question) using deterministic regex-based perturbation with fixed seeds (\texttt{RANDOM\_SEED=42}). Dollar amounts and large numbers are multiplicatively shifted by 15--40\%; percentages receive an additive shift of 3--12~pp. Injection occurs immediately after Stage 1, before any downstream agent, forcing hallucinations to propagate through the full transformation chain. All injections are logged with complete metadata enabling exact replication.

\paragraph{Detection Instruments.} We use two instruments applied independently to each of the four stage outputs. \textbf{gpt-4o Judge:} a forensic financial auditor with no ground truth access, instructed to flag suspicious claims via internal reasoning alone. \textbf{Retrieval Checker:} a fully deterministic numeric matcher (zero LLM calls) comparing agent output against FinanceBench ground truth, evidence strings, and the pre-injection researcher output at 1\% tolerance (tighter than the 2\% tolerance used in Experiment~3's RAG gate, which accommodates agent-introduced numeric reformatting such as rounding \$71.2B to \$71B). The retrieval checker represents the theoretical detection ceiling; gpt-4o Judge represents the realistic deployment scenario.

\subsection{Experiment 1: Detectability Decay Across Stages}
\label{sec:exp1}

We evaluate both detection instruments independently at all four pipeline stages across all 346 injected hallucinations (Table~\ref{tab:exp1-tab}).

\begin{figure}[h]
  \centering
  \includegraphics[width=\columnwidth]{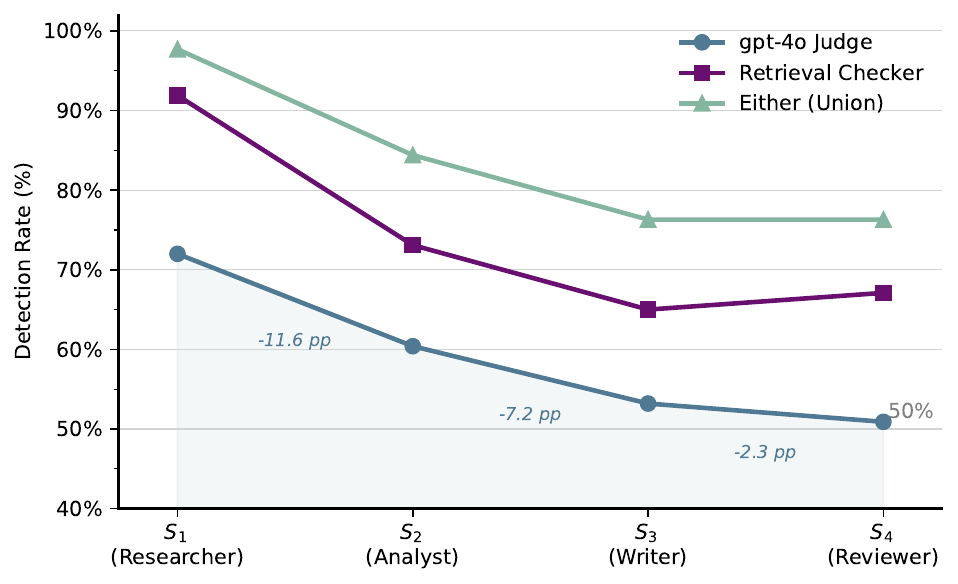}
  \caption{Hallucination detectability decays monotonically across all three detection methods ($n$=346). Every hallucination is detectable at $S_1$; gpt-4o Judge drops 21.1~pp by $S_4$, confirming the pipeline, not the hallucination itself, creates invisibility.}
  \label{fig:detection_decay}
\end{figure}

\begin{table}[h]
\centering
\resizebox{\columnwidth}{!}{%
\begin{tabular}{>{\raggedright\arraybackslash}p{3.2cm} ccc}
\toprule
\rowcolor{headerblue}
\textbf{Stage} & \textbf{gpt-4o Judge} & \textbf{Retrieval} & \textbf{Either} \\
\midrule
$S_1$ - Researcher & 72.0\% & 91.9\% & 97.7\% \\[2pt]
$S_2$ - Analyst    & 60.4\% & 73.1\% & 84.4\% \\[2pt]
$S_3$ - Writer     & 53.2\% & 65.0\% & 76.3\% \\[2pt]
$S_4$ - Reviewer   & 50.9\% & 67.1\% & 76.3\% \\[2pt]
\midrule
\textcolor{accentblue}{\textbf{Drop ($S_1{\to}S_4$)}}
  & \textcolor{accentblue}{\textbf{$-$21.1~pp}}
  & \textcolor{accentblue}{\textbf{$-$24.8~pp}}
  & \textcolor{accentblue}{\textbf{$-$21.4~pp}} \\
\bottomrule
\end{tabular}}
\vspace{5pt}
\caption{Detection rates by stage. Drop measured $S_1{\to}S_4$.}
\label{tab:exp1-tab}
\end{table}

gpt-4o detection drops 21.1~pp from Stage 1 to Stage 4. Critically, 82/346 hallucinations (23.7\%) survive completely undetected by either instrument in the final output. Zero hallucinations follow the all-missed trajectory: every injected hallucination is detectable at Stage 1. The pipeline, not the hallucination itself, is what produces invisibility. Figure~\ref{fig:detection_decay} visualizes detectability decay across all three detection methods.

Decay is boundary-concentrated, not uniform. The largest drop occurs at $S_1{\to}S_2$ (gpt-4o: $-$11.6~pp; Retrieval: $-$18.8~pp), where raw numeric facts become derived computations requiring inverse calculation to verify. A second significant drop occurs at $S_2{\to}S_3$ ($-$7.2~pp; $-$8.1~pp) as figures become causally embedded narrative. The $S_3{\to}S_4$ boundary is essentially inert ($-$2.3~pp; $+$2.1~pp): the Reviewer is editorial, not transformative. Laundering happens upstream, not at the final gate. Trajectory analysis confirms systematic behavior: 10.7\% of hallucinations follow $\checkmark{\to}\times{\to}\times{\to}\times$ (detectable at Stage 1, permanently lost thereafter) and 9.0\% follow $\checkmark{\to}\checkmark{\to}\times{\to}\times$, together accounting for 23.4\% of all cases (see Appendix~\ref{app:stats}).

\subsection{Experiment 2: Capability Ceilings of LLM Skepticism}
\label{sec:exp2}

We test whether a sufficiently capable model can overcome the structural detection ceiling (Table~\ref{tab:exp2-tab}). Four state-of-the-art models are evaluated as skeptic agents at Stage 1, the best-case position where hallucinations are still raw numerical values, on all 346 hallucinations under identical prompts and matching protocols.

\begin{table}[h]
\centering
\resizebox{\columnwidth}{!}{%
\begin{tabular}{>{\raggedright\arraybackslash}p{3.2cm} ccc}
\toprule
\rowcolor{headerblue}
\textbf{Model} & \textbf{Detection} & \textbf{95\% CI} & \textbf{Survival} \\
\midrule
Meta-Llama-3-70B-Instruct & 51.4\% & {[}46.2--56.6\%{]} & 48.6\% \\[2pt]
gemini-2.5-flash     & 72.5\% & {[}67.6--77.2\%{]} & 27.5\% \\[2pt]
gpt-4o (Exp.~1 ref.) & 72.0\% & {[}67.1--76.6\%{]} & 28.0\% \\[2pt]
DeepSeek-V3.2        & 75.7\% & {[}71.1--80.3\%{]} & 24.3\% \\[2pt]
Qwen3.5-397B-A17B         & 87.0\% & {[}83.2--90.5\%{]} & 13.0\% \\
\bottomrule
\end{tabular}}
\vspace{5pt}
\caption{Stage 1 detection rates with 95\% bootstrap CIs (5,000 resamples).}
\label{tab:exp2-tab}
\end{table}

Bootstrap CIs reveal three statistically distinct clusters: Meta-Llama-3-70B-Instruct alone at the bottom ($p < 0.000001$ vs.\ all others via McNemar tests); gemini-2.5-flash, DeepSeek-V3.2, and gpt-4o mutually indistinguishable; Qwen3.5-397B-A17B significantly above all others ($p < 0.001$ for all pairwise comparisons), yet its upper CI bound is 90.5\%. No model approaches 100\% (full pairwise tests in Appendix~\ref{app:stats}).

The ceiling is structural, not a model quality problem. Percentage hallucinations are detected at 71--98\% because a 30\%+ additive shift is often obviously implausible; dollar amounts are detected at only 49--86\% because a 12.8\% perturbation on Boeing's COGS is entirely within industry-plausible range. Plausibility reasoning cannot catch magnitude-plausible-but-wrong values without ground truth access. Even projecting the strongest model (Qwen3.5-397B-A17B, 87.0\%) through measured decay rates, Stage~4 detection falls to 60--65\%, leaving 35--40\% undetected in the final report (Appendix~\ref{app:stats}).

\subsection{Experiment 3: Timing Dominates Method}
\label{sec:exp3}

We isolate the effect of verification placement using a full gemini-2.5-flash pipeline with three conditions: \textbf{Vanilla} (no verification), \textbf{End-Check} (deterministic numeric gate after Agent 4 only, detection but no correction possible), and \textbf{Ours} (identical deterministic gates after every agent handoff, with annotated corrections passed downstream); results in Table~\ref{tab:exp3-tab}. This design isolates placement as the sole variable: same tools, different timing.

\begin{table}[h]
\centering
\resizebox{\columnwidth}{!}{%
\begin{tabular}{>{\raggedright\arraybackslash}p{3.8cm} ccc}
\toprule
\rowcolor{headerblue}
\textbf{Metric} & \textbf{Vanilla} & \textbf{End-Check} & \textbf{Ours} \\
\midrule
Survival Rate              & 60.7\% & 58.4\% & \textcolor{accentblue}{\textbf{16.2\%}} \\[2pt]
Actionable Detection       &  0.0\% &  0.0\% & \textcolor{accentblue}{\textbf{76.9\%}} \\[2pt]
Hallucination-Free Reports & 18.6\% & 20.0\% & \textcolor{accentblue}{\textbf{68.6\%}} \\[2pt]
Correction Rate            & 26.6\% & 26.3\% & \textcolor{accentblue}{\textbf{54.0\%}} \\[2pt]
Output Quality (1--5)      &  4.16  &  4.44  & \textcolor{accentblue}{\textbf{3.93}}   \\
\bottomrule
\end{tabular}}
\vspace{5pt}
\caption{Hallucination survival and key metrics ($n$=346 hallucinations, $n$=140 questions).}
\label{tab:exp3-tab}
\end{table}

\begin{figure}[h]
  \centering
  \includegraphics[width=\columnwidth]{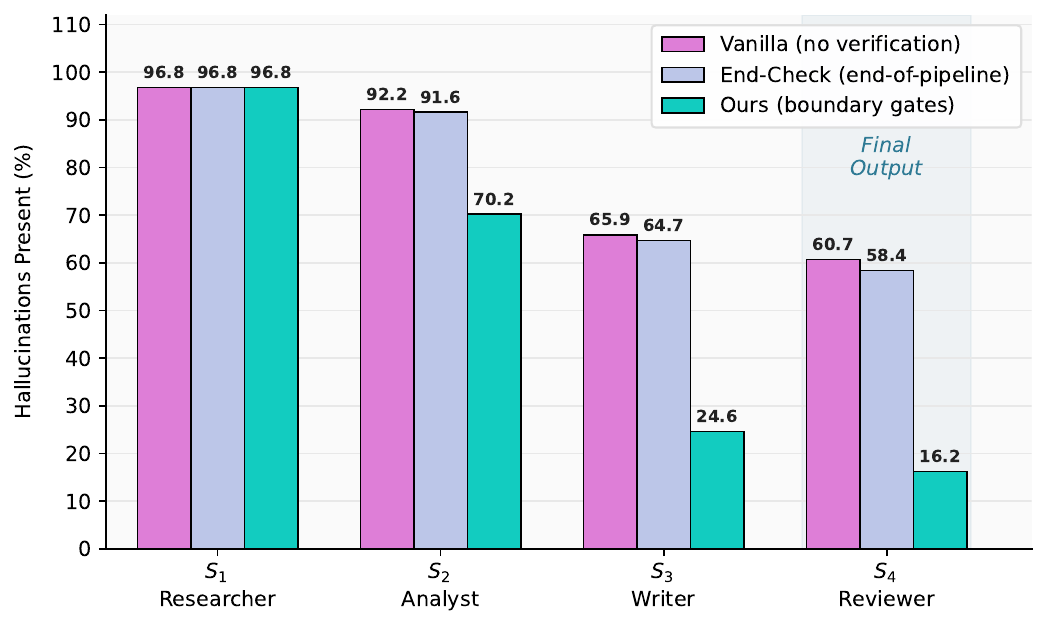}
  \caption{Boundary gates intercept the cascade at its origin while baselines decay passively. All methods begin identically at $S_1$ (96.8\%); Gate~1 alone drives a 26.6~pp drop at $S_2$, reducing final survival to 16.2\% versus 60.7\% for Vanilla, a 3.75$\times$ reduction.}
  \label{fig:propagation}
\end{figure}

End-of-pipeline checking achieves 2.3~pp improvement over no verification, statistically negligible and architecturally unactionable since no downstream agent can receive a correction. Boundary gating achieves $-$44.5~pp versus Vanilla and $-$42.2~pp versus End-Check, confirmed by five independent statistical tests (McNemar $\chi^2$=122.24, permutation $p < 0.000001$, Cohen's $h$=$-$0.911, $\chi^2$=129.94, Fisher OR=0.138; all $p < 0.000001$). Figure~\ref{fig:propagation} traces hallucination presence at each stage, confirming that boundary gates intercept the cascade at its origin.

Gate breakdown exposes the mechanism (Table~\ref{tab:gates}): Gate 1 alone captures 75.4\%, nearly as much as all three gates combined (76.9\%). Gate 3 adds only 1.5~pp of marginal coverage. Stage-by-stage presence curves confirm early interception: all three methods begin identically at $S_1$ (96.8\% presence;\footnote{The 3.2\% gap from 100\% reflects injections whose logged numeric scale differs from their text representation (e.g., \texttt{71200} vs.\ \$71.2B); the 2\% evaluation tolerance cannot bridge this unit mismatch.}); divergence begins at $S_2$ (Ours: 70.2\% vs.\ Vanilla: 92.2\%) and accelerates through $S_3$ (Ours: 24.6\% vs.\ Vanilla: 65.9\%).

\begin{table}[h]
\centering
\resizebox{\columnwidth}{!}{%
\begin{tabular}{>{\raggedright\arraybackslash}p{2.0cm}
                >{\raggedright\arraybackslash}p{2.0cm} cc}
\toprule
\rowcolor{headerblue}
\textbf{Gate} & \textbf{Position} & \textbf{Detection} & \textbf{95\% CI} \\
\midrule
Gate 1   & $S_1{\to}S_2$ & \textcolor{accentblue}{\textbf{75.4\%}} & {[}71.1--80.1\%{]} \\[2pt]
Gate 2   & $S_2{\to}S_3$ & 51.7\% & {[}46.2--57.2\%{]} \\[2pt]
Gate 3   & $S_3{\to}S_4$ & 10.7\% & {[}7.5--13.9\%{]}  \\[2pt]
\midrule
\textcolor{accentblue}{\textbf{Any Gate}}
  & ---
  & \textcolor{accentblue}{\textbf{76.9\%}}
  & \textcolor{accentblue}{\textbf{{[}72.3--81.2\%{]}}} \\
\bottomrule
\end{tabular}}
\vspace{5pt}
\caption{Per-gate detection rates (Ours only).}
\label{tab:gates}
\end{table}

The $-$0.51 quality drop (Ours vs.\ End-Check, 3.93 vs.\ 4.44) is real but its source is precise: post-hoc analysis of all 266 gate interventions finds a false positive rate of 0.4\% (1/266) - the gate almost never refutes a reference-correct value. The quality penalty instead originates from suppression-without-restoration: in 32.7\% of caught cases (87/266), the gate correctly suppressed the injected value but the downstream agent failed to propagate the annotated correction, leaving a numerical gap. This is an instruction-following limitation, not a gate accuracy problem, and points directly to source-grounded correction (injecting the correct value explicitly into agent context) as the fix. The gate precision is high; the correction propagation is not. Despite this, 96/140 hallucination-free reports versus 28/140 represents a categorical improvement in factual accuracy. A coherent report built on a fabricated figure is not a high-quality report.

\section{State-Transition Model}
\label{sec:model}

We formalize hallucination propagation as a first-order Markov process over four states defined by the form a hallucination takes at each 
pipeline stage:

$$S_1 \xrightarrow{s_1} S_2 \xrightarrow{s_2} S_3 
\xrightarrow{s_3} S_4$$

where $s_k$ is the per-boundary escape probability. States are defined by irreversible transformation: once a raw fact ($S_1$) is embedded in a derived computation ($S_2$), the original checkable claim is structurally destroyed. Downstream agents cannot recover it because they receive only transformed text. This irreversibility property explains why end-of-pipeline verification fails: the information required to verify the original claim no longer exists in verifiable form.

\begin{table}[h]
\centering
\resizebox{\columnwidth}{!}{%
\begin{tabular}{>{\raggedright\arraybackslash}p{2.0cm}
                >{\raggedright\arraybackslash}p{1.6cm}
                >{\raggedright\arraybackslash}p{2.4cm}
                >{\raggedright\arraybackslash}p{4.5cm}}
\toprule
\rowcolor{headerblue}
\textbf{Transition} & \textbf{Escape $s_k$} & \textbf{95\% CI} & \textbf{Interpretation} \\
\midrule
$S_1{\to}S_2$ & 24.6\% & {[}19.9--28.9\%{]} &
    Raw facts directly matchable; highest-value gate \\[2pt]
$S_2{\to}S_3$ & 48.3\% & {[}42.8--53.8\%{]} &
    Derived computations partially reversible \\[2pt]
$S_3{\to}S_4$ & 89.3\% & {[}86.1--92.5\%{]} & Narrative embedding is near-irreversible \\
\bottomrule
\end{tabular}}
\vspace{5pt}
\caption{Measured per-boundary escape probabilities (derived from Experiment~3 gate detection rates).}
\label{tab:markov}
\end{table}

\paragraph{Measured Transition Probabilities.} We derive escape probabilities directly from Experiment~3 gate detection rates ($s_k = 1 - \text{gate detection}_k$), shown in Table~\ref{tab:markov}.

The progression $24.6\% \to 48.3\% \to 89.3\%$ quantifies progressive irreversibility: each stage roughly doubles the fraction of hallucinations that will survive to the final output. The near-unity $S_3{\to}S_4$ escape probability (89.3\%) is the mathematical explanation for why end-checking fails: by the time an end gate runs, nearly 90\% of narrative-embedded hallucinations are structurally unrecoverable.

\paragraph{Predictive Formula.} For a linear $n$-agent pipeline with 
per-boundary escape probabilities $s_1, \ldots, s_{n-1}$:

$$P(\text{survival}) = \prod_{k=1}^{n-1} s_k$$

For our 4-agent setup: $0.246 \times 0.483 \times 0.893 \approx 10.6\%$, versus measured 16.2\%. The 5.6~pp gap reflects gate false negatives from value reformatting (e.g., ``\$71.2 billion'' $\to$ ``71.2B''), rounding, and unit conversion. The Markov model describes transformation dynamics; measured detection probabilities provide a lower bound on survival.

\paragraph{Design Prescriptions.} Three actionable prescriptions follow directly from the model, addressing FMAI's call for verified fixes with explicit resource tradeoffs. \textbf{(1) Gate placement:} invest at $S_1{\to}S_2$ first (75.4\% detection), then $S_2{\to}S_3$ (51.7\%). The $S_3{\to}S_4$ gate contributes only 10.7\% marginal detection and is economically unjustified in resource-constrained deployments. \textbf{(2) The first gate dominates:} every hallucination surviving Gate~1 faces a 48.3\% chance of surviving Gate~2 and an 89.3\% chance of surviving Gate~3. Gate~1 is worth more than Gates~2 and~3 combined. \textbf{(3) Pipeline length risk:} in a 6-agent pipeline without gates, vanilla survival would exceed our measured 60.7\% as additional transformation stages increase laundering opportunities. With gates, the model predicts survival of approximately 8--12\% depending on per-boundary escape rates, suggesting diminishing but positive returns from extending the gated architecture.

\section{Discussion}

\paragraph{Limitations.} Three limitations bound our current claims. First, our experiments operate in the financial domain, chosen precisely for its unambiguous numeric ground truth. Detection rates and decay slopes may differ in domains where facts are less crisply verifiable, such as medical summarization or legal review. The transformation mechanism ($S_1{\to}S_4$) is domain-agnostic; the specific rates are not. Second, we evaluate a strictly linear 4-agent topology. Branching, parallel, and cyclic architectures have different propagation dynamics that the current Markov formulation does not capture. Extended transition matrices for non-linear topologies remain future work. Third, our perturbation range of 15--40\% models realistic LLM hallucinations that are plausible-but-wrong. Subtle hallucinations below 5\% perturbation, which may be more common in practice for certain domains, are not tested and would likely produce lower gate detection rates given the 2\% matching tolerance. Fourth, we do not evaluate structured-handoff baselines (typed JSON fields, citation passing, or provenance-preserving programmatic state~\cite{caveagent}), which may independently reduce laundering by avoiding lossy free-text transformations entirely. Where boundary gates intercept hallucinations after transformation has begun, structured handoffs would prevent the transformation step itself. Whether this architectural alternative achieves comparable or superior survival reduction, and at what cost to pipeline flexibility, is a direct avenue for future work.

\paragraph{The Quality Metric is Structurally Blind to the Intervention's Primary Benefit.} The $-$0.51 quality drop (3.93 vs.\ 4.44) requires careful interpretation. Our evaluator scores internal consistency without access to ground truth, which means it is structurally incapable of rewarding the primary benefit of boundary gating: factual accuracy. A pipeline that produces a fluent, internally consistent report built on a fabricated \$71.2B figure scores higher than one that correctly suppresses that figure but leaves a numerical gap. This is precisely the wrong incentive structure for high-stakes deployment, and it means our quality scores systematically underestimate the true benefit of our intervention. The correct evaluation instrument is a metric that jointly scores factual accuracy and narrative coherence - for example, a retrieval-grounded rubric that checks each numerical claim in the final report against the source filing before scoring analytical quality. We expect such an instrument would not only close the observed quality gap but reverse it: a report with 96/140 hallucination-free outputs is categorically more valuable than one with 28/140, regardless of how fluent the hallucinated reports appear. Designing and validating this joint metric is a direct avenue for future work and a prerequisite for fair evaluation of any boundary-gating system. This tradeoff directly addresses FMAI's call for explicit evidence about what improves and what does not: our method improves factual accuracy substantially and measurably; the quality cost is real but measured by an instrument that cannot see the benefit it trades against.

\paragraph{Annotation-Based Correction vs.\ Source Grounding.} Our boundary gates annotate refuted values and instruct downstream agents to use corrected figures. Each gate is fully deterministic (zero LLM calls); in our setup, boundary gating adds approximately 0.9 seconds of total overhead across three gates due to rate limiting, negligible relative to the 3--5 second per-agent Gemini latency. This produces a 22.9~pp gap between actionable detection (76.9\%) and full correction (54.0\%): the hallucination is suppressed but the correct value is not always propagated. A stronger intervention would retrieve and inject the correct value directly into agent context rather than relying on instruction-following to propagate an annotation. We expect source grounded correction to close this gap substantially and reduce the quality penalty simultaneously, since agents would have the correct value explicitly available rather than inferring it from an annotation.

\paragraph{Generalization to Other Pipeline Topologies.} The Markov formulation extends naturally to $n$-agent linear pipelines via the product formula $P(\text{survival}) = \prod_{k=1}^{n-1} s_k$. For non-linear topologies, the key insight generalizes even if the formula does not: verification value is highest immediately before a transformation that destroys verifiability, and lowest after narrative embedding has occurred. In branching pipelines, this prescribes gates at every merge point where outputs from parallel agents are synthesized. In memory-augmented pipelines, it prescribes verification before any memory write, since a hallucination written to persistent memory propagates to every future agent that reads it.

\paragraph{Why This Matters Beyond Finance.} The hallucination snowball is not a financial analysis problem. It is a systems architecture problem that manifests wherever sequential agents transform each other's outputs without structured verification. Medical summarization pipelines that chain extraction, synthesis, and report generation face identical dynamics. Legal review pipelines that chain document parsing, precedent retrieval, and brief drafting face identical dynamics. Any domain where a wrong fact can be transformed into a confident narrative and then institutionally approved faces this failure mode. The measurement infrastructure and gate architecture we provide are domain-agnostic in mechanism; practitioners in other domains should expect different per-boundary escape rates requiring domain-specific calibration.

\section{Conclusion}

We present the hallucination snowball effect: a formally characterized, empirically measured, and theoretically modeled failure mode in sequential multi-agent LLM pipelines. Hallucinations do not merely persist across agent handoffs. They transform through predictable stages that progressively destroy verifiability, from a checkable raw fact at $S_1$ to an editorially approved conclusion at $S_4$ that no downstream agent can recover. gpt-4o detection degrades 21.1~pp across four stages. 23.7\% of hallucinations escape entirely. Even the strongest model tested faces a structural ceiling of 87.0\% at best-case conditions, projected to 35--40\% survival at Stage~4.

The solution is not a better detector. It is earlier intervention. Boundary gates using identical tools to end-of-pipeline checking achieve a 42.2~pp reduction in hallucination survival (Cohen's $h = -0.911$), confirmed by five independent statistical tests. The Markov model explains precisely why: the $S_3{\to}S_4$ escape probability is 89.3\%, meaning that by the time an end gate runs, nearly 90\% of narrative-embedded hallucinations are already unrecoverable. Gate~1 alone, placed at $S_1{\to}S_2$, captures 75.4\% of all hallucinations while they are still raw numerical facts and directly matchable against reference values.

This intervention comes with a measured cost: output quality drops from 4.44 to 3.93 on internal consistency scoring, reflecting a genuine accuracy-coherence tradeoff. Post-hoc analysis attributes this primarily to suppression-without-restoration (32.7\% of caught cases) rather than false positives (0.4\%), pointing to source-grounded correction as the direct fix. We report this honestly because FMAI's call for explicit evidence about what improves and what does not demands it. The tradeoff is real, but it is preferable to the alternative: a fluent, coherent report built entirely on fabricated figures.

The reframe this paper offers is simple and actionable: from ``can we detect hallucinations?'' to ``can we control their propagation?'' For practitioners deploying multi-agent pipelines in high-stakes domains today, this is not a theoretical concern. It is a measurable, reproducible, and now-addressable systems engineering problem. Act before transformation, not after.

\newpage


\bibliography{example_paper}
\bibliographystyle{icml2026}

\clearpage
\appendix

\section{Full Statistical Results}
\label{app:stats}

\subsection{Hallucination Trajectory Distribution (Experiment 1, 
\textit{n}=346)}

\begin{table}[h]
\centering
\label{tab:trajectories}
\resizebox{\columnwidth}{!}{%
\begin{tabular}{>{\raggedright\arraybackslash}p{4.2cm} cc
                >{\raggedright\arraybackslash}p{3.5cm}}
\toprule
\rowcolor{headerblue}
\textbf{Trajectory} & \textbf{Count} & \textbf{\%} &
\textbf{Interpretation} \\
\midrule
$\checkmark{\to}\checkmark{\to}\checkmark{\to}\checkmark$
    & 244 & 70.5\% & Always detected \\[2pt]
$\checkmark{\to}\times{\to}\times{\to}\times$
    & 37  & 10.7\% & Full decay at $S_1{\to}S_2$; permanently lost \\[2pt]
$\checkmark{\to}\checkmark{\to}\times{\to}\times$
    & 31  &  9.0\% & Decay at $S_2{\to}S_3$ \\[2pt]
$\checkmark{\to}\times{\to}\checkmark{\to}\times$
    & 11  &  3.2\% & Intermittent \\[2pt]
$\checkmark{\to}\checkmark{\to}\times{\to}\checkmark$
    &  8  &  2.3\% & Late recovery \\[2pt]
$\times{\to}\checkmark{\to}\checkmark{\to}\checkmark$
    &  4  &  1.2\% & Late detection \\[2pt]
$\checkmark{\to}\times{\to}\times{\to}\checkmark$
    &  4  &  1.2\% & Partial recovery \\[2pt]
$\times{\to}\times{\to}\checkmark{\to}\checkmark$
    &  2  &  0.6\% & Very late detection \\[2pt]
$\checkmark{\to}\checkmark{\to}\checkmark{\to}\times$
    &  2  &  0.6\% & Final-stage laundering \\[2pt]
$\times{\to}\times{\to}\times{\to}\checkmark$
    &  1  &  0.3\% & Single-stage late detection \\[2pt]
$\times{\to}\times{\to}\times{\to}\times$
    & 0 & 0.0\% & Never occurs \\
\bottomrule
\end{tabular}}
\vspace{5pt}
\caption{Hallucination trajectory distribution (Experiment~1, $n$=346). The $\times{\to}\times{\to}\times{\to}\times$ count is exactly zero: every hallucination is detectable at $S_1$.}
\end{table}

The all-missed trajectory count is exactly zero: every injected hallucination is detectable at Stage 1 by at least one instrument. The pipeline creates invisibility; the hallucination does not arrive invisible.

\subsection{Experiment 2: All Six Pairwise McNemar Tests (\textit{n}=346)}

\begin{table}[h]
\centering
\label{tab:mcnemar}
\resizebox{\columnwidth}{!}{%
\begin{tabular}{>{\raggedright\arraybackslash}p{3.0cm} cccc}
\toprule
\rowcolor{headerblue}
\textbf{Pair} & \textbf{$\Delta$ (pp)} & \textbf{$\chi^2$} &
\textbf{$p$-value} & \textbf{Cohen's $h$} \\
\midrule
Meta-Llama-3-70B-Instruct vs.\ gemini-2.5-flash    & $-$21.1 &  44.31 & $<$0.000001 & $-$0.439 \\[2pt]
Meta-Llama-3-70B-Instruct vs.\ DeepSeek-V3.2  & $-$24.3 &  63.79 & $<$0.000001 & $-$0.511 \\[2pt]
Meta-Llama-3-70B-Instruct vs.\ Qwen3.5-397B-A17B      & $-$35.5 & 108.64 & $<$0.000001 & $-$0.804 \\[2pt]
gemini-2.5-flash vs.\ DeepSeek-V3.2 & $-$3.2  &   1.12 & 0.289       & $-$0.073 \\[2pt]
gemini-2.5-flash vs.\ Qwen3.5-397B-A17B     & $-$14.5 &  30.01 & $<$0.000001 & $-$0.365 \\[2pt]
DeepSeek-V3.2 vs.\ Qwen3.5-397B-A17B   & $-$11.3 &  18.28 & 0.000019    & $-$0.293 \\
\bottomrule
\end{tabular}}
\vspace{5pt}
\caption{All six pairwise McNemar tests.}
\end{table}

McNemar contingency tables (both\_caught / only\_A\_caught / 
only\_B\_caught / both\_missed): Meta-Llama-3-70B-Instruct vs.\ gemini-2.5-flash: 156/22/95/73. 
Meta-Llama-3-70B-Instruct vs.\ DeepSeek-V3.2: 166/12/96/72. Meta-Llama-3-70B-Instruct vs.\ Qwen3.5-397B-A17B: 171/7/130/38. 
gemini-2.5-flash vs.\ DeepSeek-V3.2: 212/39/50/45. gemini-2.5-flash vs.\ Qwen3.5-397B-A17B: 236/15/65/30. 
DeepSeek-V3.2 vs.\ Qwen3.5-397B-A17B: 242/20/59/25.

\subsection{Experiment 3: Five Independent Statistical Tests}

\textit{Ours vs.\ Vanilla} (16.2\% vs.\ 60.7\%, $\Delta = -44.5$~pp): McNemar $\chi^2 = 136.10$ ($p < 0.000001$); permutation test $p < 0.000001$ (10,000 permutations); Cohen's $h = -0.958$ (large effect); unpaired $\chi^2 = 142.95$ ($p < 0.000001$); Fisher's exact OR $= 7.996$ ($p < 0.000001$). McNemar contingency: both\_survived $= 47$, only\_Vanilla\_survived $= 9$, only\_Ours\_survived $= 163$, both\_caught $= 127$.

\textit{Ours vs.\ End-Check} (16.2\% vs.\ 58.4\%, $\Delta = -42.2$~pp): McNemar $\chi^2 = 122.24$ ($p < 0.000001$); permutation test $p < 0.000001$ (10,000 permutations); Cohen's $h = -0.911$ (large effect); unpaired $\chi^2 = 129.94$ ($p < 0.000001$); Fisher's exact OR $= 0.138$ ($p < 0.000001$). McNemar contingency: both\_survived $= 43$, only\_End-Check\_survived $= 13$, only\_Ours\_survived $= 159$, both\_caught $= 131$. All five tests are mutually consistent; $p$-values are orders of magnitude below any Bonferroni-corrected threshold.

\subsection{Experiment 2: Detection by Injection Type}

\begin{table}[h]
\centering
\label{tab:injtype}
\resizebox{\columnwidth}{!}{%
\begin{tabular}{>{\raggedright\arraybackslash}p{3.2cm} ccc}
\toprule
\rowcolor{headerblue}
\textbf{Model} & \textbf{Dollar Amount} &
\textbf{Large Number} & \textbf{Percentage} \\
\midrule
Meta-Llama-3-70B-Instruct
    & 49.3\% (133/270) & 37.0\% (10/27) & 71.4\% (35/49) \\[2pt]
gemini-2.5-flash
    & 69.6\% (188/270) & 55.6\% (15/27) & 98.0\% (48/49) \\[2pt]
DeepSeek-V3.2
    & 71.1\% (192/270) & 85.2\% (23/27) & 95.9\% (47/49) \\[2pt]
Qwen3.5-397B-A17B
    & 85.6\% (231/270) & 81.5\% (22/27) & 98.0\% (48/49) \\
\bottomrule
\end{tabular}}
\vspace{5pt}
\caption{Detection rates by injection type at Stage~1, Experiment~2. Percentage hallucinations are easiest to catch; dollar amounts are hardest, explaining the structural ceiling.}
\end{table}

This asymmetry directly explains the structural ceiling: percentage hallucinations are caught at 71--98\% because a large additive shift is often obviously implausible; dollar amount hallucinations are caught at only 49--86\% because a 12.8\% perturbation on a large company's financials is entirely within industry-plausible range without ground truth access.

\section{Injection Protocol}

\subsection{Regex Patterns}
Three numeric expression classes are detected and perturbed, applied case-insensitively in priority order.

\begin{itemize}[leftmargin=*]
  \item \textbf{Dollar amounts:} patterns beginning with \texttt{\$} followed by comma-separated digits, optional decimal, and optional scale suffixes (million, billion, mn, bn, k, MM, trillion, etc.)
  \item \textbf{Percentages:} signed or unsigned numeric values followed by \texttt{\%}, \texttt{percent}, \texttt{percentage points}, or \texttt{bps}
  \item \textbf{Large numbers:} non-dollar, non-percentage sequences of five or more digits with optional decimals
\end{itemize}

Year exclusion: values in range 1900--2099 are excluded when surrounding context (20 chars before, 10 chars after) contains any of: FY, FISCAL, YEAR, Q1--Q4, 10-K, 10-Q, ANNUAL, QUARTER. Any 4-digit value in 1950--2030 is treated as a year unconditionally.

\subsection{Perturbation Rules and Seed Structure}
Dollar amounts and large numbers are multiplicatively shifted by 15--40\% of their original value; percentages receive an additive shift of 3--12 percentage points. Direction (increase or decrease) is chosen randomly per injection. Each question receives 2--3 injections. Seeds are fully deterministic: \texttt{RANDOM\_SEED=42}; primary seed per question $=$ \texttt{RANDOM\_SEED $+$ abs(hash(question\_id)) $\%$ 10000}; per-injection sub-seed $=$ \texttt{primary\_seed $+$ abs(hash(question\_id)) $\%$ 100000 $+$ i $\times$ 7919}. Retry on no-change: re-attempts with sub-seed $+$ 999983.

\section{Infrastructure}

Frameworks: Python, LangGraph (StateGraph). APIs: OpenAI (gpt-4o, gpt-4o-mini), Google GenAI SDK (gemini-2.5-flash), HuggingFace Router / Novita (Meta-Llama-3-70B-Instruct, DeepSeek-V3.2, Qwen3.5-397B-A17B). Rate limiting: 0.3-second delay between API calls (Experiment~3). Retry logic: 3 attempts with 2-second base delay across all experiments.

\end{document}